\documentclass[conference]{IEEEtran}
\IEEEoverridecommandlockouts
\usepackage{amsmath,amssymb}  
\PassOptionsToPackage{dvipsnames,table}{xcolor}
\usepackage{xcolor}
\usepackage{graphicx}
\usepackage{multirow}
\usepackage{algorithmic}
\usepackage{cite}

\usepackage{amsfonts}
\usepackage{textcomp}
\usepackage{epsfig}
\usepackage[normalem]{ulem}
\useunder{\uline}{\ul}{}
\usepackage{color}
\usepackage{booktabs}
\usepackage{stfloats}
\usepackage{booktabs}
\usepackage{colortbl}
\usepackage{epsfig}
\usepackage{hhline}
\usepackage{arydshln} 
\usepackage{pifont}
\usepackage{enumitem}
\usepackage{colortbl}
\usepackage{makecell}
\usepackage{verbatim}
\usepackage{bm}
\usepackage{caption,subcaption}
\usepackage{bbding}

\def\BibTeX{{\rm B\kern-.05em{\sc i\kern-.025em b}\kern-.08em
    T\kern-.1667em\lower.7ex\hbox{E}\kern-.125emX}}
\begin{document}

\title{Adaptive Low Light Enhancement via Joint Global-Local Illumination Adjustment

%\thanks{Identify applicable funding agency here. If none, delete this.}
}

\author{\IEEEauthorblockN{1\textsuperscript{st} Haodian Wang}
\IEEEauthorblockA{\textit{University of Science and Technology of China} \\
Hefei, Anhui \\
wanghaodian@mail.ustc.edu.cn}
\and
\IEEEauthorblockN{2\textsuperscript{nd} Yaqi Song}
\IEEEauthorblockA{\textit{Chn Energy Digital Intelligence Technology} \\ \textit{Development (Beijing) CO., LTD.} \\
syqdgmzxx@gmail.com}
}

\maketitle

\begin{abstract}
Images captured under real-world low-light conditions face significant challenges due to uneven ambient lighting, making it difficult for existing end-to-end methods to enhance images with a large dynamic range to normal exposure levels. To address the above issue, we propose a novel brightness-adaptive enhancement framework designed to tackle the challenge of local exposure inconsistencies in real-world low-light images. Specifically, our proposed framework comprises two components: the Local Contrast Enhancement Network (LCEN) and the Global Illumination Guidance Network (GIGN). We introduce an early stopping mechanism in the LCEN and design a local discriminative module, which adaptively perceives the contrast of different areas in the image to control the premature termination of the enhancement process for patches with varying exposure levels. Additionally, within the GIGN, we design a global attention guidance module that effectively models global illumination by capturing long-range dependencies and contextual information within the image, which guides the local contrast enhancement network to significantly improve brightness across different regions. Finally, in order to coordinate the LCEN and GIGN, we design a novel training strategy to facilitate the training process. Experiments on multiple datasets demonstrate that our method achieves superior quantitative and qualitative results compared to state-of-the-art algorithms. The source codes will be publicly available once the paper is accepted.
% The LCEN 
%We introduce an early stopping mechanism to terminate the enhancement process for regions with sufficient exposure. Additionally, we design a global attention guidance module that effectively models global illumination by capturing long-range dependencies and contextual information within the image, which guides the local contrast enhancement network to significantly improve brightness across different regions. Finally, in order to coordinate LCEN and GIGN, we design a novel training strategy to facilitate the training procedure. Experiments on multiple datasets demonstrate that our method achieves better quantitative and qualitative results than the state-of-the-art algorithms. The source codes will be publicly available once the paper is accepted.
% 真实场景的低光照环境下捕捉到的图像面临一系列挑战，例如不均匀的环境光照。由于同一幅图像中存在的光照水平不同，大多数现有的方法直接对其进行增强，这导致了不同区域间的曝光不一致问题。

%由于实际的弱光场景中可能面临着不均匀的环境光照，导致图像中存在较大的动态范围，因此现有的端到端方法很难将不均匀低光图像增强到正常状态。为了解决上述问题，我们将低光图像增强视作全局引导的局部照明增强任务，并提出了一个亮度自适应增强框架，以处理现实情况下弱光图像中局部曝光不一致的问题。具体而言，我们的增强框架由两部分构成，即局部对比度增强网络和全局照明引导网络。我们引入了早停机制，对于曝光充分的区域，提前退出增强过程。同时，我们设计了全局注意力引导网模块，通过捕获图像中的远程依赖性和上下文信息对全局光照进行有效建模，从而引导局部增强网络有效提升不同区域的亮度水平。为了进一步验证所提出方法的有效性，。

\end{abstract}

\begin{IEEEkeywords}
Low-light image enhancement, Global-local illumination adjustment, Uneven exposure correction.
\end{IEEEkeywords}

\section{Introduction}
\begin{figure}
    \centering
    \includegraphics[width=1\linewidth]{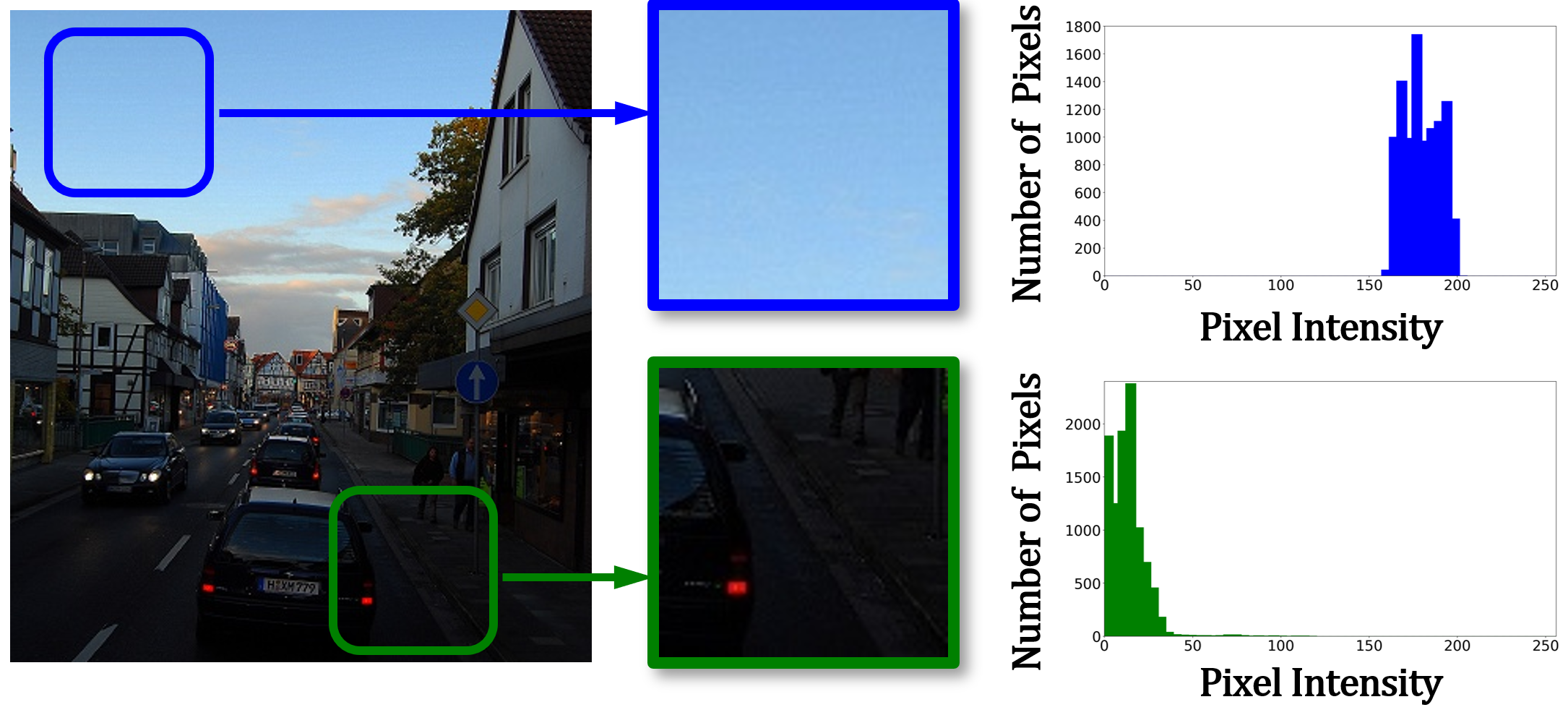}
    \caption{Histogram of brightness distribution in different regions of low-light images from natural scenes.}
    \label{fig:fig1}
    \vspace{-5.0mm}
\end{figure}

Images captured under low-light conditions often suffer from exposure inconsistencies due to uneven light distribution and varying object reflectance \cite{survey}. The phenomenon significantly impairs both human visual perception and the performance of advanced visual algorithms \cite{object,tracking}. Consequently, enhancing low-light images captured in scenes with a wide dynamic range to achieve normal exposure has garnered significant attention from researchers.
% 在真实场景下捕捉的低光照图像可能会存在场景亮度分布不均匀问题。由于拍摄角度、场景背光的原因
As shown in Fig. \ref{fig:fig1}, we demonstrate the inconsistency in the brightness distribution across different regions of the same low-light image. Therefore, applying a globally consistent enhancement across different regions may struggle to effectively represent local brightness. Moreover, enhancing areas with lower brightness is considerably more challenging than enhancing regions with sufficient illumination because of the substantial variations in brightness distribution.
% 在低照度下捕获的图像会遭受统计和结构特性失真，导致低对比度、模糊和噪声问题。这会显着降低人类视觉效果和高级视觉算法的性能。由于真实场景中亮度的高动态范围，不同亮度下的对比度下降和细节纹理失真是不同的。因此，如何将宽动态范围低光场景下拍摄的低光图像增强至正常曝光，受到了研究人员的极大关注。
% 如图1所示，我们展示了在同一张低光照图像上的不同区域，其亮度分布的不一致性。因此，使用同一个网络对其进行全局增强可能难以有效表征局部亮度，且由于亮度差异较大，亮度较低的区域会相对于亮度较为充足的区域增强部分即可。

%在这篇文章中，我们提出了一种新颖的自适应亮度增强框架，以解决现实场景中低光图像中的局部曝光不一致问题。我们设计了带有局部判别模块的局部对比度增强网络，输入图像以分而治之的方式被切分成小块分别进行增强，并通过局部鉴别模块判断是否达到良好光照水平。同时我们引入了早停机制，以自适应控制其提前退出增强过程，从而解决输入低光照图像局部对比度不一致性的问题，同时避免过度增强。此外，我们设计了一个全局照明引导网络来感知照明，它可以有效地捕获图像内的远程依赖性和全局上下文信息，辅助局部增强网络提升增强结果的质量。最后我们设计了一种新颖的训练策略来有效地约束所提出框架的优化过程。综合实验表明，我们提出的方法在混合低光数据集中实现了最先进的性能。
%为了解决真实场景下拍摄的图像存在的曝光不一致问题，我们设计了一个新颖的局部增强网络。具体而言，

% 通过联合全局照明来引导局部对比度增强，该框架可以自适应地应用变化的(2)提出了局部判别模块

In this paper, we propose a novel adaptive brightness enhancement framework to address the challenge of local exposure inconsistencies in low-light images from real-world environments. We design a Local Contrast Enhancement Network (LCEN) integrated with a Local Discriminative Module (LDM), where input image patches with varying illumination intensities are enhanced through distinct network pathways. The LDM evaluates whether the illumination level of each patch is sufficient. Additionally, we incorporate an early stopping mechanism \cite{earlystop} to adaptively regulate the termination of the enhancement process, which addresses the problem of local contrast inconsistency in low-light images while avoiding over-enhancement. Furthermore, we design a Global Illumination Guidance Network (GIGN) to perceive illumination, which effectively captures long-range dependencies and global contextual information within the image, and assists the local enhancement network in improving the quality of the enhanced results. Finally, we propose a novel training strategy to effectively constrain the optimization process of the proposed framework. Comprehensive experiments demonstrate that the proposed method achieves state-of-the-art performance on diverse low-light datasets.

The contributions can be summarized as follows:

(1) We propose a novel adaptive brightness enhancement framework to address local exposure inconsistencies in low-light images from real-world scenarios. By incorporating an early stopping mechanism and utilizing global illumination to guide local contrast enhancement, the framework can adaptively apply varying degrees of enhancement based on the differing brightness of regions. 
%为了有效解决真实场景的低光照图像中存在的曝光水平不一致问题，我们提出了一种新颖的自适应亮度增强框架，通过联合全局信息引导局部增强，同时设置了早退机制，能够使得图像中曝光水平不一致的区域进行不同程度的增强。
%受到动态神经网络启发
%将不均匀低光图像增强问题视为自适应增强过程，我们提出了一种新颖的递归增强框架，通过感知亮度分布并平衡细节和对比度增强来实现自适应图像增强。将不均匀采用分而治之的策略 设计了全局局部网络，并训练策略
%提出了一种新颖的 ACT-Net 来同时增强图像对比度和结构，引入 CDC 来平衡细节和亮度增强。此外，为了自适应增强在不同光照水平下捕获的图像，提出了一种新颖的 BP-Net 来感知亮度以控制 ACT-Net 的递归增强次数。

(2) A Local Discriminative Module is proposed to adaptively control the Local Contrast Enhancement Network for improving the local contrast of the image. Additionally, we design a Global Illumination Guidance Network to perceive illumination, which effectively captures long-range dependencies and global contextual information within the image. 

(3) We design a novel training strategy to effectively constrain the optimization process of the proposed framework.

(4) Comprehensive experiments demonstrate that, compared with the eleven low-light image enhancement methods, our proposed method achieves state-of-the-art performance on diverse low-light datasets.

\begin{figure*}
\setlength{\belowcaptionskip}{-0.3cm}
    \centering
    \includegraphics[width=0.98\linewidth]{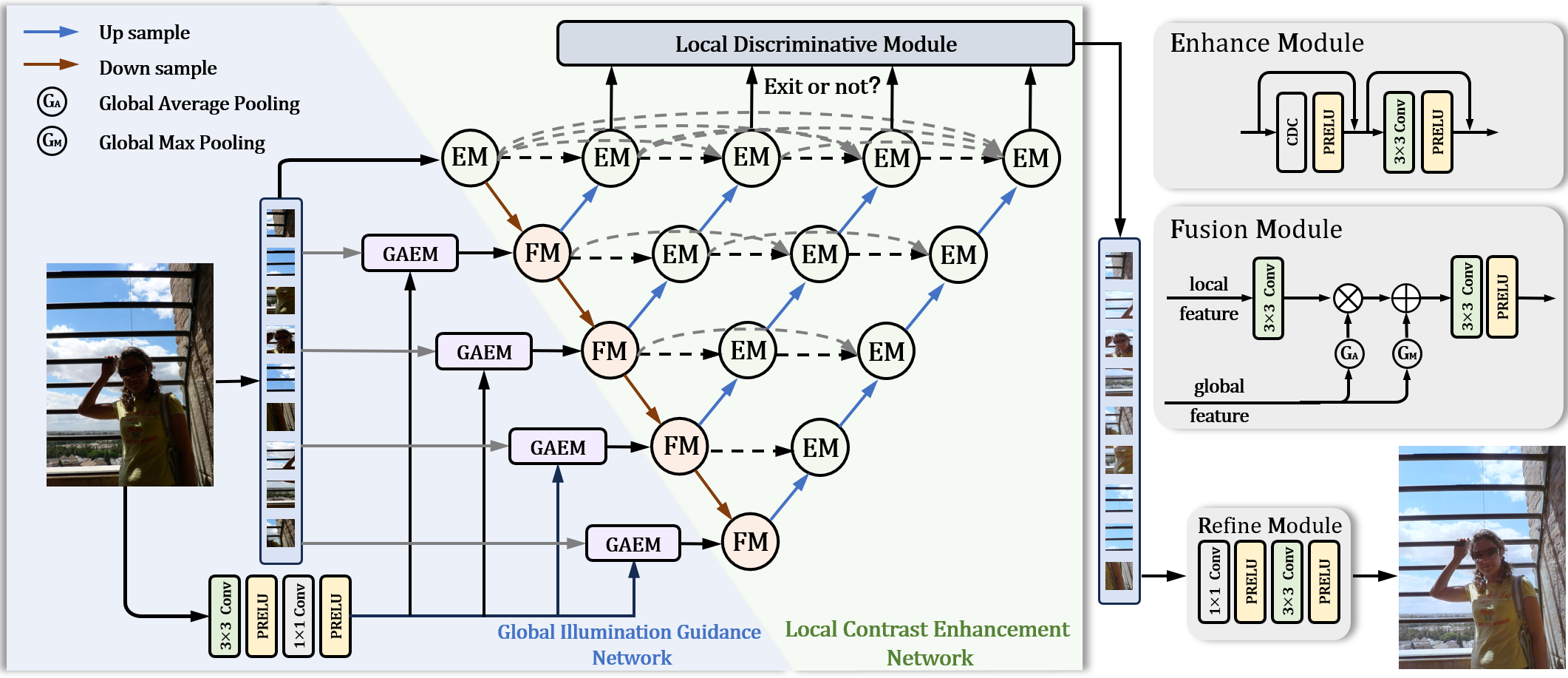}
    \caption{The overall architecture of proposed enhancement framework, which consists of the Local Contrast Enhancement Network (LCEN) and Global Illumination Guidance Network (GIGN).}
    \label{fig:fig2}
\end{figure*}

\section{Related Work}
Low-light image enhancement methods can be approximately categorized as traditional methods and learning-based methods \cite{sharma2021nighttime,myMMM,myMMM2,myTAI,RUAS,RetinexNet&LOL,DRBN,DCE,SCI}. Traditional methods encompass techniques such as histogram equalization \cite{HE}, curve mapping \cite{LIME}, and Retinex theory \cite{retinex}. However, these approaches are highly reliant on handcrafted image priors and are thus limited in their ability to handle complex real-world scenarios.

With the rapid advancement of deep learning, many learning-based methods have been successively proposed, achieving impressive performance in the LLIE. Zhou et al. \cite{GLARE} proposed a Low-Light Image Enhancement (LLIE) network named GLARE, which augments low-light images via codeword retrieval of generated latent features. Li et al. \cite{COTF} proposed a real-time exposure correction method named Collaborative Transformation Framework, which efficiently integrates global transformations with pixel-level transformations. Dang et al. \cite{ppformer} proposes a lightweight CNN-transformer hybrid network using pixel-wise and patch-wise cross-attention mechanisms for low-light image enhancement. However, these methods primarily mitigate either overexposure or underexposure in input images, and they still struggle to effectively enhance images with uneven illumination.

\begin{figure}
    \centering
    \includegraphics[width=0.8\linewidth]{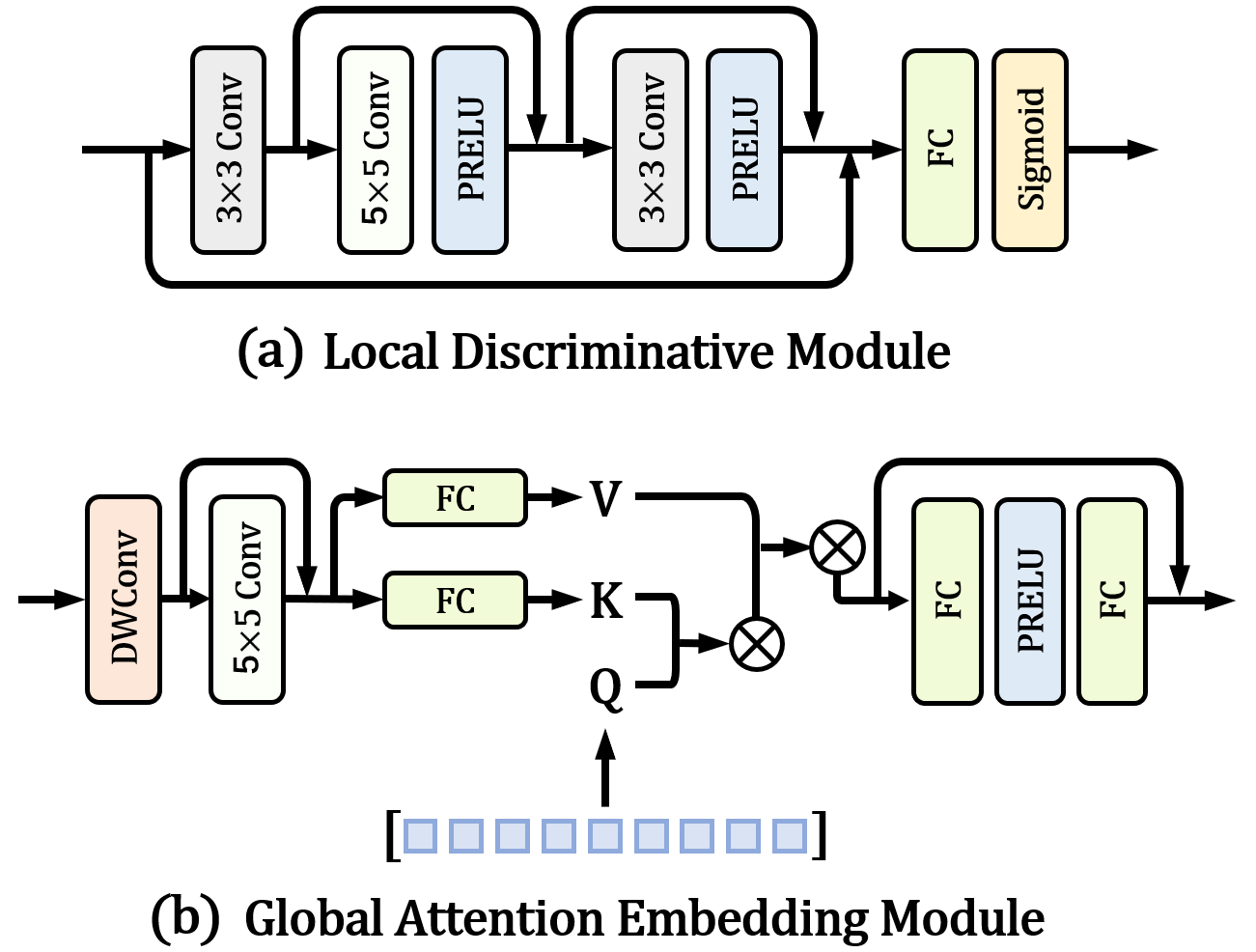}
    \caption{The detailed architecture of Local Discriminative Module (LDM) and Global Attention Embedding Module (GAEM).}
    \label{fig:fig3}
    \vspace{-5.0mm}
\end{figure}
\section{Global-Local Illumination Adjustment Network}
In this section, we introduce the architecture of the proposed Global-Local Illumination Adjustment Network (GLIAN). The holistic network comprises three main components: the Global Illumination Guidance Network (GIGN), the Local Contrast Enhancement Network (LCEN), and Refine Module (RM). As illustrated in the Fig. \ref{fig:fig2}, the input image is segmented into distinct patches, which are then fed into the LCEN for adaptive enhancement. Concurrently, the original input image is processed by the GIGN to generate guidance factors that adaptively modulate the local enhancement process. Subsequently, the enhanced local patches are assembled and further fine-tuned through the refinement network to remove artifacts and enhance details. The specific details are outlined as follows.
%在本节中，我们介绍了提出的联合局部-全局照明调整网络的架构。整体网络由Global Illumination Guidance Network、Local Contrast Enhancement Network以及最后的细化模块组成，如图所示，输入图像被分割成不同的patch，送到Local Contrast Enhancement Network中进行自适应的增强；并行地，原始输入图像被送到Global Illumination Guidance Network生成引导因子，对局部增强过程进行自适应的调制。最后，增强后的局部patch被组合到一起，通过细化网络进行进一步的微调以去除伪影增强细节。具体细节如下所述。

\subsection{Local Contrast Enhancement Network}

%To tackle the challenge of inconsistent exposure in images captured in real-world environments, we design an novel Local Contrast Enhancement Network (LCEN). Specifically, the input image is divided into different patches in a divide-and-rule manner for individual enhancement. A Local Discrimination Module (LDM) is utilized to assess whether a satisfactory illumination level has been achieved, enabling adaptive control for early exit from the enhancement process. This approach resolves the problem of inconsistent local contrast in low-light input images while avoiding over-enhancement. The core components of the LEN consist of U-shaped skip-densely connected Enhancement Module (EM) blocks and Feature Modulation (FM) blocks. Each EM block is composed of cascaded 3×3 convolutions and Parametric Rectified Linear Units (PReLU). Given that enhancing contrast and reinforcing structural textures are equally crucial during the illumination enhancement process, we have incorporated a Contrast and Detail Enhancement (CDC) module to assist the network in adaptively enhancing fine-grained texture and structural information. Additionally, we have introduced Spatial Feature Transformation to design the FM module, which effectively guides and modulates local features using global attention features derived from a Global Guidance Network (GGN), thereby significantly enhancing the expressive capacity of the network.
To address the issue of inconsistent exposure in images captured under real-world scenarios, we propose a novel Local Contrast Enhancement Network. Specifically, the input image is divided into different patches following a divide-and-conquer strategy, with each patch independently enhanced. A Local Discriminative Module (LDM) is employed to assess whether a patch has achieved optimal illumination levels, enabling adaptive early termination of the enhancement process. This design effectively mitigates local contrast inconsistency in low-light images while preventing over-enhancement.

\begin{figure*}
    \centering
    \includegraphics[width=1\linewidth]{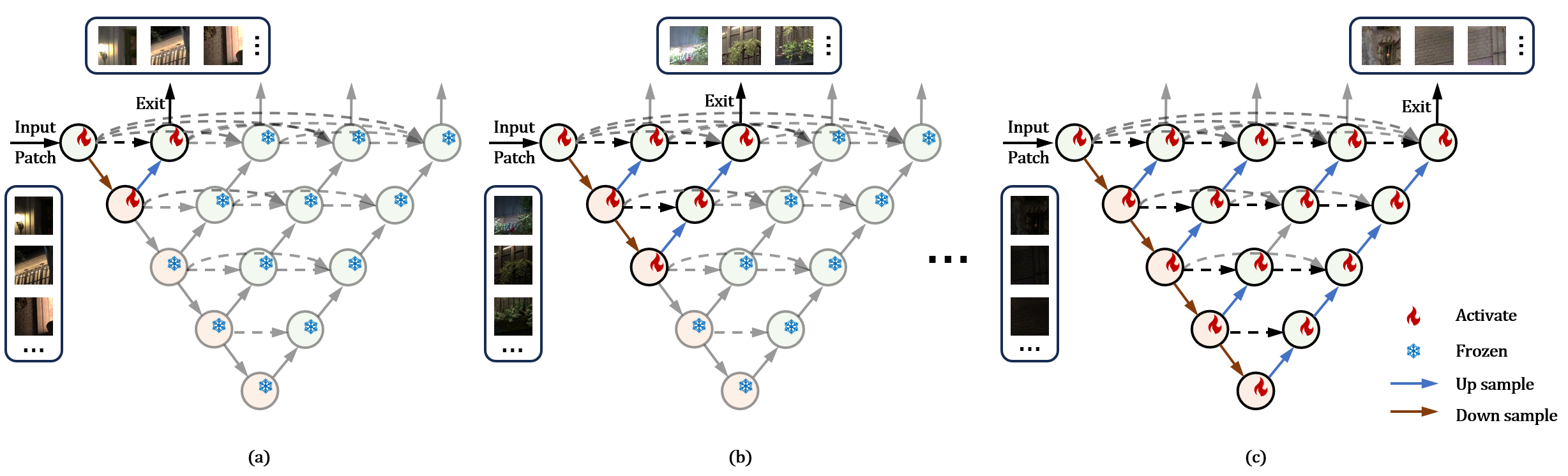}
    \caption{Training strategy of the overall framework.}
    \label{fig:fig4}
    \vspace{-5.0mm}
\end{figure*}
As shown in Fig. \ref{fig:fig3}[a], the local discriminative module consists of cascaded 3 × 3 and 5 × 5 convolution layers, a fully connected layer, and a Sigmoid activation function. Compared to patches with optimal exposure levels, those with suboptimal exposure are processed through deeper network layers to achieve more thorough enhancement. The local discriminative module then determines whether a patch has reached an adequate illumination level and controls the network's exit. The selective mechanism allows several patches to pass through fewer network layers, which may effectively reduce network redundancy and improve inference speed.

Moreover, the main architecture of the Local Enhancement Network comprises U-shaped densely connected EM blocks and FM blocks. Each EM block is constructed using cascaded 3 × 3 convolution layers and PReLU activation functions. Given that both contrast enhancement and texture structure improvement are critical in illumination enhancement, we introduce a CDC \cite{CDC} convolution module to assist the network in adaptively enhancing fine-grained texture details. Furthermore, we incorporate a spatial feature transformation mechanism \cite{lin2017inverse} within the FM block. By leveraging global guidance features $F_g$ extracted from the global illumination guidance network, the FM block effectively guides and modulates local features $F^p_l$, thereby significantly enhancing the network's expressive capability. The fusion process can be expressed as the following:
\begin{equation}
\small
F^p_f=PReLU(Conv(Conv(F^p_l) \cdot G_A(F_g)) + G_M(F_g) )),
\end{equation}
where $p$ represents different feature patches, and $F^p_f$ represents the fused output feature. $G_A$ and $G_M$ represent global average pooling and global mean pooling, respectively.
%为了解决真实场景下拍摄的图像存在的曝光不一致问题，我们设计了一个新颖的局部增强网络。具体而言，输入图像以分而治之的方式被切分成小块分别进行增强，并通过局部鉴别模块判断是否达到良好光照水平，以自适应控制其提前退出增强过程，从而解决输入低光照图像局部对比度不一致性的问题，同时避免过度增强。局部增强网络主体部分由u形跳过稠密连接的em块和fm块组成。em块由级联的3*3卷积和prelu组成.由于在光照增强的过程中提升对比度和增强结构纹理同样重要，因此我们添加了CDC模块以辅助网络自适应增强细粒度的纹理结构信息。此外，我们引入了空间特征变换设计了fm模块，通过来自全局引导网络的全局注意力特征对局部特征进行有效的引导与调制，从而有效提升网络的表达能力。融合过程如下：

%由于需要增强的程度不同，为了自适应选择不同的patch所需增强的路径，我们设计了局部鉴别模块对增强过程中的patch进行自适应选择，以确保高光区域不会过度增强，同时欠曝光区域能够得到充分增强。如图5a所示，局部鉴别模块由级联的3*3卷积、5*5卷积、全连接层和Sigmoid激活函数组成，其目的在于自适应选择patch是否达到增强标准。对于不同对比度的patch而言，如图4所示，光照水平良好的patch会相对于光照水平欠佳的patch通过更深的网络以进行更加充分的增强，最后通过局部鉴别模块鉴别是否达到良好光照水平并退出网络。通过这种选择机制，能够使得部分patch只需要通过部分网络路径就能够有效增强，有效降低了网络冗余并提高了推理速度。

%全局引导网络为了保证在局部增强的过程中，能够有效融入全局对比度信息，设计了全局引导网络。我们引入了自注意力机制，对于进行特征嵌入受到了目标检测 DETR网络的启发，我们将随机初始化的query输入到模块中，与图像自身生成的key和value共同作用，最终输出十个参数

\subsection{Global Illumination Guidance Network}

To maintain global consistency while improving local contrast and details, we introduce a self-attention mechanism to integrate global contextual information. The input image and the patches are fused through the Global Attention Embedding Module (GAEM) to generate the guiding map, after being processed through a cascade of 3 × 3 convolutions, 1 × 1 convolutions, and PReLU activation. Inspired by DETR \cite{DETR}, as illustrated in Fig. \ref{fig:fig3}[b], we feed image patches into the module as embeddings, where they interact with keys and values generated from the original image through weighted operations to produce guiding variables. By modeling contextual correlations to identify complementary cues across the entire spatial domain, we enable adaptive modulation of the local enhancement process.
%为了在增强局部对比度和细节的过程中对全局上下文信息进行建模，从而维持全局一致性，我们引入了自注意力机制来整合全局信息。输入图像通过级联的3*3卷积、1*1卷积和PRELU后，与patchs一起被馈送入GAEM中进行融合以生成引导map。受到DETR的启发，如图所示，我们将patch块转化成embedding，输入到模块中，与图像本身生成的key和value进行加权运算，从而生成引导变量。通过建模空间相关性来寻找整个空间域内的互补线索，我们能够对局部增强过程进行自适应的调制。

\subsection{Refine Module}

Given that the direct splicing of enhanced local patches may induce artifacts and grid effects, we introduce a lightweight refine module subsequent to the local enhancement process. The refinement module comprises two sets of consecutively stacked 1 × 1 convolutional layers and 3 × 3 convolutional layers, each followed by PReLU activation. By deeply integrating the features extracted from various local patches, we are able to effectively eliminate potential artifacts and generate more natural and enhanced results.

%\subsection{Loss Function}

\subsection{Training Strategy and Loss Funtion}
%w为了自适应
\textbf{(Stage \uppercase\expandafter{\romannumeral1}) Pretrain the Local Contrast Enhancement Network (excluding the Local Discriminative Module)}. We divide low-light images into non-overlapping patches and compute the average value of the brightness channel. Subsequently, these patches are classified into four levels, ranging from low to high, based on their average brightness values. 

\begin{figure*}
    \centering
    \setlength{\abovecaptionskip}{0.15cm}
\setlength{\belowcaptionskip}{0cm}   \includegraphics[width=1.0\linewidth]{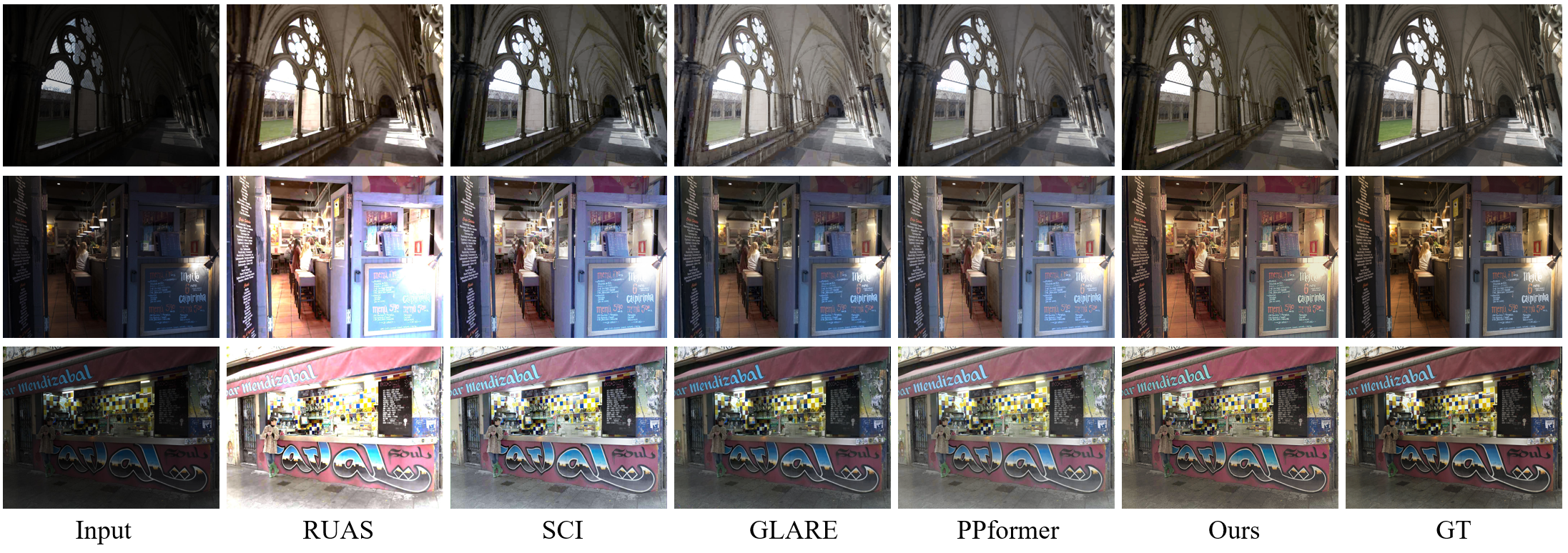}
    \caption{Qualitative comparison of results on a dataset composed of three representative datasets: LOL \cite{RetinexNet&LOL}, MIT-Adobe FiveK \cite{MIT}, and SICE \cite{SICE}.}
    \label{fig:fig5}
    \vspace{-3.0mm}
\end{figure*}

\begin{table*}[ht]
\small
\centering
\caption{Comparison with eight existing low light enhancement methods on our constructed datasets by three no-reference and three reference metrics. The best and second-best performances are marked as {\color{red} \textbf{bold}} and {\color{blue} \textbf{bold}}, respectively.}
\setlength{\tabcolsep}{5mm}{
\begin{tabular}{c|cccccc}
\Xhline{2.\arrayrulewidth}
\hline
                          & \multicolumn{6}{c}{Metrics}                                                                                                                                                                       \\ \cline{2-7} 
\multirow{-2}{*}{Methods} & PSNR↑                          & SSIM↑                         & LPIPS↓                        & LOE↓                            & NIQE↓                         & EME↑                           \\ \hline
DRBN                      & 18.6135                        & 0.7847                        & 0.1947                        & 508.33                        & 3.0890                        & 7.7540                         \\
DCE++                     & 18.3141                        & 0.7252                        & 0.1885                        & 271.76                        & 3.3478                        & 10.4613                        \\
RUAS                      & 16.5633                        & 0.6785                        & 0.2008                        & 278.67                        & 3.9528                        & 9.7951                         \\
SCI                       & 19.4286                        & 0.7365                        & 0.1905                        & 258.15                        & 3.2922                        & 12.7043                        \\
ZERO-IG                   & 19.6388                        & 0.8076                        & 0.1664                        & 307.28                        & 2.6894                        & 11.4895                        \\
CoTF                      & 20.9971                        & 0.7962                        & 0.1573                        & 251.9455                        & 2.5132                        & {\color{blue} \textbf{12.7468}}                       \\
GLARE                     & 20.9164                        & 0.8069                        & {\color{red} \textbf{0.1519}} & 302.1825                        & 2.4195                        & 10.6482                        \\
PPformer                  & {\color{blue} \textbf{21.0293}}                        & {\color{blue} \textbf{0.8104}}                        & 0.1552                        & {\color{blue} \textbf{245.95}}                        & {\color{red} \textbf{2.3965}} & {\color{red} \textbf{13.3795}} \\
Ours                      & {\color{red} \textbf{21.3724}} & {\color{red} \textbf{0.8129}} & {\color{blue} \textbf{0.1532}}                        & {\color{red} \textbf{234.96}} & {\color{blue} \textbf{2.4151}}                        & 12.1598                        \\ \hline
\Xhline{2.\arrayrulewidth}
\end{tabular}}
\label{tab:tab1}
\end{table*}

As shown in Fig. \ref{fig:fig4}(a), when training with the brightest patches, we update the backpropagation parameters only in the shallowest layers, while freezing the other layers. Conversely, when training with the darkest patches, as shown in Fig. \ref{fig:fig4}(c), we update the parameters across all layers. Overall, we constrain the network using L1 loss between the enhanced patches $p$
and ground truth $\hat{P}$:
\begin{equation}
L_{stage_{1}} = \lambda _{L_1}\cdot ||P-\hat{P}||_1.
\end{equation}

\textbf{(Stage \uppercase\expandafter{\romannumeral2}) Pretrain the Local Discriminative Module}. We utilize a batch of patches with varying contrast levels, including underexposed patches and normally exposed patches, to train a binary classification network. The network assigns the classification result as "exit" for normally exposed patches and "not exit" otherwise. In the second stage, we apply the CrossEntropyLoss function to impose the constraint:
\begin{equation}
L_{stage_2} = \lambda _{CE}\cdot L_{CE}(C,\hat{C}),
\end{equation}
where $C$ and $\hat{C}$ represent output classification results and real labels, respectively.

\textbf{(Stage \uppercase\expandafter{\romannumeral3}) Fine-tuning overall network}. After pretraining each module, we perform joint fine-tuning on the entire network. To eliminate artifacts across different patches, we introduce a Structural Similarity (SSIM) loss \cite{SSIM}. Overall, we constrain the network using L1 loss and SSIM loss between the enhanced image $I$ and ground truth $\hat{I}$:
\begin{equation}
L_{stage_3} = \lambda _{L_1}\cdot ||I-\hat{I}||_1 +  \lambda _{ssim}\cdot SSIM(I,\hat{I}).
\end{equation}

\begin{figure*}
    \centering
    \setlength{\abovecaptionskip}{0.15cm}
\setlength{\belowcaptionskip}{0cm}   \includegraphics[width=0.8\linewidth]{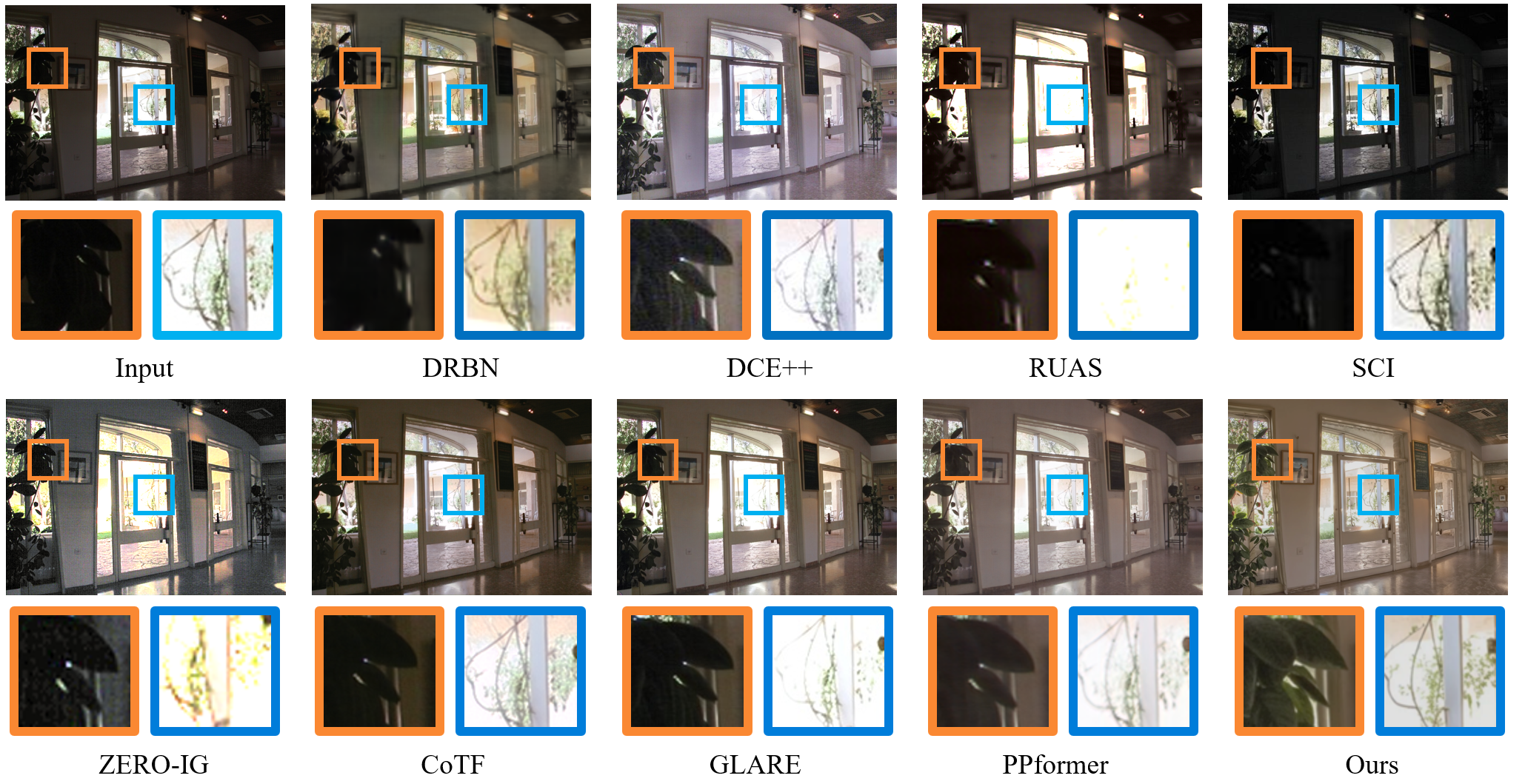}
    \caption{Detailed comparison with existing representative methods on five representative real-world datasets (including NPE, LIME, MEF, DICM, and VV).}
    \label{fig:fig6}
    \vspace{-3.0mm}
\end{figure*}
\begin{table*}[ht]
\small
\centering
\caption{A comparison of the performance of three no-reference metrics and three reference metrics for eight existing low-light enhancement methods across five representative real-world datasets (including NPE, LIME, MEF, DICM, and VV). The best result is shown in \textcolor{red}{red}, and the second-best result is \textcolor{blue}{blue}.}
\setlength{\tabcolsep}{2mm}{
\begin{tabular}{c|ccccccccc}
\Xhline{2.\arrayrulewidth}
\hline
\multirow{2}{*}{Metrics} & \multicolumn{9}{c}{Methods}                                                       \\ \cline{2-10} 
                         & DRBN   & DCE++  & RUAS   & SCI    & ZERO-IG & CoTF   & GLARE  & PPformer & Ours   \\ \hline
LOE↓                     & 673.60 & 385.37 & 541.71 & 376.92 & 362.96  & {\color{blue} \textbf{356.62}} & 377.64 & 359.73   & {\color{red} \textbf{351.92}} \\
NIQE↓                    & 3.92   & 3.53   & 4.96   & 3.84   & 3.78    & 3.47   & 3.52   & {\color{blue} \textbf{3.42}}     & {\color{red} \textbf{3.41}}   \\
EME↑                     & 5.99   & 6.49   & 5.68   & 6.63   & 6.57    & {\color{red} \textbf{6.98}}   & 6.74   & {\color{blue} \textbf{6.84}}     & 6.51   \\ \hline
\Xhline{2.\arrayrulewidth}
\end{tabular}}
\label{tab:tab2}
\end{table*}

\section{Experiments and Analysis}
\subsection{Experimental Setting}
\textbf{Datasets.}
We selected 2,500 images with varying brightness distributions from three representative public datasets (including LOL \cite{RetinexNet&LOL}, MIT-Adobe FiveK \cite{MIT}, and SICE \cite{SICE}) to conduct a series of meticulously designed experiments. Given the uniqueness of the SICE dataset, which contains a series of multi-exposure images covering a range of brightness levels, we created image pairs by selecting the darkest image from each sequence and its corresponding ground truth under normal lighting conditions. Furthermore, to assess the generalization ability of the proposed method in real-world scenarios, we conducted additional comparative experiments on five real-world datasets, including NPE \cite{wang2013naturalness}, LIME \cite{LIME}, MEF \cite{MEF}, DICM \cite{DICM}, and VV \footnote{ https://sites.google.com/site/vonikakis/datasets}.

\textbf{Training setting.}
In the experiments, we implemented the proposed framework using PyTorch  on a single NVIDIA 3090 GPU, employing the Adam optimizer with parameters $\beta_{1}$ = 0.9, $\beta_{2}$ = 0.999, and $\varepsilon = 10^{-8}$. The learning rate and batch size were set to $10^{-8}$ and 8, respectively. The total number of epochs was set to 300.

\textbf{Comparison methods and metrics.}
We compare proposed method with eight advanced low light image enhancement methods, including DRBN \cite{DRBN}, DCE$++$ \cite{DCE}, RUAS \cite{RUAS}, SCI \cite{SCI}, ZERO-IG \cite{shi2024zero}, CoTF \cite{COTF}, GLARE \cite{GLARE} and PPformer \cite{ppformer}. To comprehensively demonstrate the superiority of our method, we use three reference metrics and three no-reference metrics to evaluate the performance. For reference metrics, we use the PSNR $\uparrow$\footnote{ $\uparrow$ means the higher, the better, $\downarrow$ means the lower, the better.} , SSIM $\uparrow$ \cite{SSIM} and LPIPS $\downarrow$\footnotemark[2] \cite{LPIPS}. For no-reference metrics, we use NIQE $\downarrow$ \cite{NIQE}, EME $\uparrow$ \cite{EME} and LOE $\downarrow$ \cite{LOE} for evaluation.

\subsection{Qualitative results}
%我们在由三个代表性的配对数据集（包括LOL、SICE、MIT）混合而成的数据集上展示了对比实验的主观结果。如图6所示，RUAS方法在产生了过曝，导致局部信息丢失。此外GLARE方法在全局光照恢复中表现不佳，并难以有效同时处理不同亮度水平的照明。这主要是因为它们没有对不同区域的增强程度作细致的切分，从而降低了它们在实际不均匀低光照图像中的增强效率。此外，为了证明提出的方法的泛化能力，我们在代表性的无参考数据集上比较了现有方法的细节。如图7所示，DRBN和SCI方法难以有效增强欠曝光区域，而Zero-IG方法会导致部分区域过曝从而丢失细节。相比之下，提出的方法能够同时有效恢复不同亮度水平的照明。
We present the subjective results of comparative experiments on a dataset composed of three representative paired datasets, including LOL, SICE, and MIT. As shown in Fig. \ref{fig:fig5}, the RUAS method causes overexposure, leading to a loss of local information. Additionally, the GLARE method performs poorly in global illumination recovery and struggles to effectively handle lighting across different brightness levels simultaneously. This is mainly due to their lack of fine-grained segmentation of enhancement degrees for different regions, which reduces their enhancement efficiency on real-world uneven low-light images. Furthermore, to demonstrate the generalization ability of the proposed method, we compare existing methods on representative no-reference datasets. As shown in Fig. \ref{fig:fig6}, the DRBN and SCI methods fail to effectively enhance underexposed regions, while the Zero-IG method causes overexposure in some areas, resulting in the loss of details. In contrast, the proposed method is capable of simultaneously and effectively recovering illumination across different brightness levels.

\subsection{Quantitative results}
%如表1所示，我们在不同方法之间进行了定量比较。我们利用了三个全参考指标和三个无参考指标。提出的方法在大多数指标中取得了卓越的结果，特别超过了PSNR，SSIM和NIQE上现有的LLIE方法。这些结果表明，提出的方法在照明增强方面具有优势。
As shown in Table \ref{tab:tab1} and Table \ref{tab:tab2}, we conducted a quantitative comparison among different methods using three full-reference metrics and three no-reference metrics. The proposed method achieved outstanding results across most of the metrics, particularly surpassing existing LLIE methods in PSNR, SSIM, and LOE. These results demonstrate the superior performance of the proposed method in both global and local illumination enhancement.

\subsection{Computational cost comparison}
To further demonstrate the advantages of our proposed method, we conducted a detailed comparison of computational costs with existing methods. The results are presented in Table \ref{tab:tab3}. It is evident that our method not only achieves state-of-the-art performance, but also significantly reduces computational costs compared to some large-scale low-light image enhancement models like GLARE\cite{GLARE}.
\begin{table*}[ht]
\small
\centering
\caption{Comparison with existing methods on computational cost. The best performance and second best performance are marked as {\color[HTML]{FF0000} \textbf{bold}} and {\color{blue} \textbf{bold}}, respectively.}
\setlength{\tabcolsep}{3mm}{

\begin{tabular}{c|ccccc}
\Xhline{2.\arrayrulewidth}
\hline
Methods  & FLOPs(G) & Params(M) & Time(S)  & PSNR↑   & SSIM↑  \\ \hline
DRBN     & 37.7902  & 0.577168  & 0.065768 & 18.6135 & 0.7847 \\
DCE++    & 5.2112   & 0.078912  & 0.002375 & 18.3141 & 0.7252 \\
RUAS     & 0.2813   & 0.001437  & 0.042659 & 16.5633 & 0.6785 \\
SCI      & 0.062    & 0.000348  & 0.001688 & 19.4286 & 0.7365 \\
ZERO-IG  & 11.8725  & 0.123628  & 0.150337 & 19.6388 & 0.8076 \\
CoTF     & 1.8162   & 0.319454   & 0.009552   & 20.9971 & 0.7962 \\
GLARE    & 17.6213  & 0.622352   & 0.421225   & 20.9164 & 0.8069 \\
PPformer & 3.7125   & 0.095134   & 0.037179   & {\color{blue} \textbf{21.0293}} & {\color{blue} \textbf{0.8104}} \\
Ours     & 3.9562   & 0.076359    & 0.046231  & {\color{red} \textbf{21.3724}} & {\color{red} \textbf{0.8129}} \\ \hline
\Xhline{2.\arrayrulewidth}
\end{tabular}}
\label{tab:tab3}
\end{table*}

\begin{table}[ht]
\footnotesize
\caption{Ablation on different frameworks and modules. The best results are highlighted in {\color{red} \textbf{red}}. LDM means local discriminative module, GIGN means global illumination guidance network, GAEM means global attention embedding module, RM means refine module.}
\setlength{\tabcolsep}{3mm}{
\begin{tabular}{cccc|cc}
\hline
\Xhline{2.\arrayrulewidth}
\multicolumn{4}{c|}{ablation study setting} & \multicolumn{2}{c}{performance} \\ \hline
w/LDM     & w/GIGN    & w/GAEM    & w/RM    & PSNR            & SSIM          \\ \hline
×         & $\surd $         & $\surd $         & $\surd $       & 17.63           & 0.76          \\
$\surd $         & $\surd $         & ×         & ×       & 19.87           & 0.80          \\
$\surd $         & $\surd $         & $\surd $         & ×       & 20.04           & 0.81          \\
$\surd $         & $\surd $         & $\surd $         & $\surd $       & {\color{red} \textbf{20.13}}           & {\color{red} \textbf{0.82}}          \\ \hline
\Xhline{2.\arrayrulewidth}
\end{tabular}}
\label{tab:tab4}
\end{table}
\subsection{Ablation study}
\begin{figure}
    \centering
    \setlength{\abovecaptionskip}{0.15cm}
\setlength{\belowcaptionskip}{0cm}   \includegraphics[width=1.0\linewidth]{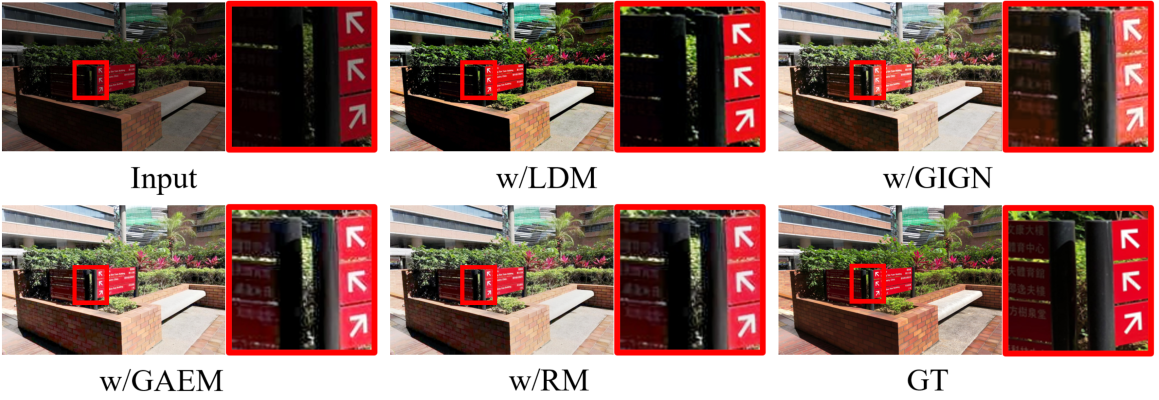}
    \caption{Ablation study on different modules.}
    \label{fig:fig7}
    \vspace{-3.0mm}
\end{figure}

To validate the effectiveness of proposed modules, We conducted ablation studies respectively.
%局部鉴别模块，全局引导模块，细化模块
%\ subsubsection {局部增强机制的效果}
%为了验证所提出的局部增强机制的有效性，我们去除了局部鉴别模块，对于输入图像的所有patch，均通过完整的网络。 如图8和表2所示，对不同亮度的区域使用相同的网络深度可能会导致t图像的全局照明均被增强，导致部分区域过曝，从而导致各种性能度量的下降。 
%\ subsubsection {所有模块的有效性} 
%为了验证各个模块的有效性，我们首先只保留局部增强模块，随后在网络中逐步添加了使用相等数量的香草卷积层代替了提出的全局注意力嵌入模块的全局照明引导网络、全局注意力嵌入模块和细化模块。实验结果如图8和表2所示，去除全局照明引导网络或全局注意力嵌入模块会导致对比度增强有效性的降低。此外，缺乏细化模块可能会导致图像的边缘纹理模糊。
%\subsubsection{Effect of Local Enhancement Module}
To validate the effectiveness of the proposed local enhancement mechanism, we removed the Local Discriminative Module and processed all patches of the input image through the complete network. As shown in Fig. \ref{fig:fig7} and Table \ref{tab:tab4}, using the same network depth for regions with different brightness levels may result in global illumination enhancement across the image, leading to overexposure in certain areas, which in turn causes a decline in various performance metrics.

%\subsubsection{Contribution of Each Module}
To verify the effectiveness of other module, we first retained only the local enhancement module and then gradually added the global illumination guidance network, the global attention embedding module, and the refinement module, each replaced by an equal number of vanilla convolutional layers. The experimental results, as shown in Fig. \ref{fig:fig7} and Table \ref{tab:tab4}, indicate that removing the global illumination guidance network or the global attention embedding module leads to a decrease in the effectiveness of contrast enhancement. Additionally, the absence of the refinement module may result in blurred edge textures in the image.

\section{Conclusion}
In this paper, we propose a brightness-adaptive enhancement framework to address the challenges posed by uneven low-light images with wide dynamic ranges in real-world scenarios. Specifically, Our framework comprises two key components: the Local Contrast Enhancement Network (LCEN) and the Global Illumination Guidance Network (GIGN). We also incorporate an early stopping mechanism, that adaptively perceives the contrast of different regions in the image, to control the enhancement process. Additionally, we propose a global attention guidance module that models global illumination by capturing long-range dependencies and contextual information, thereby guiding the LCEN to significantly enhance brightness across diverse regions. Furthermore, we design a novel training strategy to facilitate the coordination between the LCEN and GIGN. Experimental results on multiple datasets demonstrate that our method achieves superior quantitative and qualitative performance compared to state-of-the-art algorithms.

%\section*{Acknowledgment}

%\section*{References}
\bibliographystyle{IEEEtran}
\small\bibliography{main}

% Generated by IEEEtran.bst, version: 1.14 (2015/08/26)
\begin{thebibliography}{10}
\providecommand{\url}[1]{#1}
\csname url@samestyle\endcsname
\providecommand{\newblock}{\relax}
\providecommand{\bibinfo}[2]{#2}
\providecommand{\BIBentrySTDinterwordspacing}{\spaceskip=0pt\relax}
\providecommand{\BIBentryALTinterwordstretchfactor}{4}
\providecommand{\BIBentryALTinterwordspacing}{\spaceskip=\fontdimen2\font plus
\BIBentryALTinterwordstretchfactor\fontdimen3\font minus
  \fontdimen4\font\relax}
\providecommand{\BIBforeignlanguage}[2]{{%
\expandafter\ifx\csname l@#1\endcsname\relax
\typeout{** WARNING: IEEEtran.bst: No hyphenation pattern has been}%
\typeout{** loaded for the language `#1'. Using the pattern for}%
\typeout{** the default language instead.}%
\else
\language=\csname l@#1\endcsname
\fi
#2}}
\providecommand{\BIBdecl}{\relax}
\BIBdecl

\bibitem{survey}
C.~Li, C.~Guo, L.-H. Han, J.~Jiang, M.-M. Cheng, J.~Gu, and C.~C. Loy,
  ``Low-light image and video enhancement using deep learning: A survey,''
  \emph{IEEE Transactions on Pattern Analysis and Machine Intelligence}, pp.
  1--1, 2021.

\bibitem{object}
T.-Y. Lin, P.~Doll{\'a}r, R.~Girshick, K.~He, B.~Hariharan, and S.~Belongie,
  ``Feature pyramid networks for object detection,'' in \emph{Proceedings of
  the IEEE conference on computer vision and pattern recognition}, 2017, pp.
  2117--2125.

\bibitem{tracking}
Q.~Chu, W.~Ouyang, H.~Li, X.~Wang, B.~Liu, and N.~Yu, ``Online multi-object
  tracking using cnn-based single object tracker with spatial-temporal
  attention mechanism,'' in \emph{Proceedings of the IEEE international
  conference on computer vision}, 2017, pp. 4836--4845.

\bibitem{earlystop}
T.~Bolukbasi, J.~Wang, O.~Dekel, and V.~Saligrama, ``Adaptive neural networks
  for efficient inference,'' in \emph{International Conference on Machine
  Learning}.\hskip 1em plus 0.5em minus 0.4em\relax PMLR, 2017, pp. 527--536.

\bibitem{sharma2021nighttime}
A.~Sharma and R.~T. Tan, ``Nighttime visibility enhancement by increasing the
  dynamic range and suppression of light effects,'' in \emph{Proceedings of the
  IEEE/CVF Conference on Computer Vision and Pattern Recognition}, 2021, pp.
  11\,977--11\,986.

\bibitem{myMMM}
H.~Wang, Y.~Wang, Y.~Cao, and Z.-J. Zha, ``Fusion-based low-light image
  enhancement,'' in \emph{International Conference on Multimedia
  Modeling}.\hskip 1em plus 0.5em minus 0.4em\relax Springer, 2023, pp.
  121--133.

\bibitem{myMMM2}
H.~Wang, ``Frequency-based unsupervised low-light image enhancement
  framework,'' in \emph{International Conference on Multimedia Modeling}.\hskip
  1em plus 0.5em minus 0.4em\relax Springer, 2025, pp. 427--439.

\bibitem{myTAI}
H.~Wang, L.~Peng, Y.~Sun, Z.~Wan, Y.~Wang, and Y.~Cao, ``Brightness perceiving
  for recursive low-light image enhancement,'' \emph{IEEE Transactions on
  Artificial Intelligence}, 2023.

\bibitem{RUAS}
R.~Liu, L.~Ma, J.~Zhang, X.~Fan, and Z.~Luo, ``Retinex-inspired unrolling with
  cooperative prior architecture search for low-light image enhancement,'' in
  \emph{Proceedings of the IEEE/CVF Conference on Computer Vision and Pattern
  Recognition}, 2021, pp. 10\,561--10\,570.

\bibitem{RetinexNet&LOL}
C.~Wei, W.~Wang, W.~Yang, and J.~Liu, ``Deep retinex decomposition for
  low-light enhancement,'' \emph{British Machine Vision Conference}, 2018.

\bibitem{DRBN}
W.~Yang, S.~Wang, Y.~Fang, Y.~Wang, and J.~Liu, ``From fidelity to perceptual
  quality: A semi-supervised approach for low-light image enhancement,'' in
  \emph{Proceedings of the IEEE/CVF conference on computer vision and pattern
  recognition}, 2020, pp. 3063--3072.

\bibitem{DCE}
C.~Li, C.~Guo, and C.~C. Loy, ``Learning to enhance low-light image via
  zero-reference deep curve estimation,'' \emph{IEEE transactions on pattern
  analysis and machine intelligence}, vol.~44, no.~8, pp. 4225--4238, 2021.

\bibitem{SCI}
L.~Ma, T.~Ma, R.~Liu, X.~Fan, and Z.~Luo, ``Toward fast, flexible, and robust
  low-light image enhancement,'' in \emph{Proceedings of the IEEE/CVF
  Conference on Computer Vision and Pattern Recognition}, 2022, pp. 5637--5646.

\bibitem{HE}
S.~M. Pizer, E.~P. Amburn, J.~D. Austin, R.~Cromartie, A.~Geselowitz, T.~Greer,
  B.~ter Haar~Romeny, J.~B. Zimmerman, and K.~Zuiderveld, ``Adaptive histogram
  equalization and its variations,'' \emph{Computer vision, graphics, and image
  processing}, vol.~39, no.~3, pp. 355--368, 1987.

\bibitem{LIME}
X.~Guo, Y.~Li, and H.~Ling, ``Lime: Low-light image enhancement via
  illumination map estimation,'' \emph{IEEE Transactions on image processing},
  vol.~26, no.~2, pp. 982--993, 2016.

\bibitem{retinex}
E.~H. Land, ``The retinex theory of color vision,'' \emph{Scientific american},
  vol. 237, no.~6, pp. 108--129, 1977.

\bibitem{GLARE}
H.~Zhou, W.~Dong, X.~Liu, S.~Liu, X.~Min, G.~Zhai, and J.~Chen, ``Glare: Low
  light image enhancement via generative latent feature based codebook
  retrieval,'' in \emph{European Conference on Computer Vision}.\hskip 1em plus
  0.5em minus 0.4em\relax Springer, 2025, pp. 36--54.

\bibitem{COTF}
Z.~Li, F.~Zhang, M.~Cao, J.~Zhang, Y.~Shao, Y.~Wang, and N.~Sang, ``Real-time
  exposure correction via collaborative transformations and adaptive
  sampling,'' in \emph{Proceedings of the IEEE/CVF Conference on Computer
  Vision and Pattern Recognition}, 2024, pp. 2984--2994.

\bibitem{ppformer}
J.~Dang, Y.~Zhong, and X.~Qin, ``Ppformer: Using pixel-wise and patch-wise
  cross-attention for low-light image enhancement,'' \emph{Computer Vision and
  Image Understanding}, vol. 241, p. 103930, 2024.

\bibitem{CDC}
Z.~Yu, C.~Zhao, Z.~Wang, Y.~Qin, Z.~Su, X.~Li, F.~Zhou, and G.~Zhao,
  ``Searching central difference convolutional networks for face
  anti-spoofing,'' in \emph{Proceedings of the IEEE/CVF conference on computer
  vision and pattern recognition}, 2020, pp. 5295--5305.

\bibitem{lin2017inverse}
C.-H. Lin and S.~Lucey, ``Inverse compositional spatial transformer networks,''
  in \emph{Proceedings of the IEEE conference on computer vision and pattern
  recognition}, 2017, pp. 2568--2576.

\bibitem{DETR}
X.~Zhu, W.~Su, L.~Lu, B.~Li, X.~Wang, and J.~Dai, ``Deformable detr: Deformable
  transformers for end-to-end object detection,'' \emph{arXiv preprint
  arXiv:2010.04159}, 2020.

\bibitem{MIT}
V.~Bychkovsky, S.~Paris, E.~Chan, and F.~Durand, ``Learning photographic global
  tonal adjustment with a database of input/output image pairs,'' in \emph{CVPR
  2011}.\hskip 1em plus 0.5em minus 0.4em\relax IEEE, 2011, pp. 97--104.

\bibitem{SICE}
J.~Cai, S.~Gu, and L.~Zhang, ``Learning a deep single image contrast enhancer
  from multi-exposure images,'' \emph{IEEE Transactions on Image Processing},
  vol.~27, no.~4, pp. 2049--2062, 2018.

\bibitem{SSIM}
Z.~Wang, A.~C. Bovik, H.~R. Sheikh, and E.~P. Simoncelli, ``Image quality
  assessment: from error visibility to structural similarity,'' \emph{IEEE
  transactions on image processing}, vol.~13, no.~4, pp. 600--612, 2004.

\bibitem{wang2013naturalness}
S.~Wang, J.~Zheng, H.-M. Hu, and B.~Li, ``Naturalness preserved enhancement
  algorithm for non-uniform illumination images,'' \emph{IEEE transactions on
  image processing}, vol.~22, no.~9, pp. 3538--3548, 2013.

\bibitem{MEF}
K.~Ma, K.~Zeng, and Z.~Wang, ``Perceptual quality assessment for multi-exposure
  image fusion,'' \emph{IEEE Transactions on Image Processing}, vol.~24,
  no.~11, pp. 3345--3356, 2015.

\bibitem{DICM}
C.~Lee, C.~Lee, and C.-S. Kim, ``Contrast enhancement based on layered
  difference representation,'' in \emph{2012 19th IEEE international conference
  on image processing}.\hskip 1em plus 0.5em minus 0.4em\relax IEEE, 2012, pp.
  965--968.

\bibitem{shi2024zero}
Y.~Shi, D.~Liu, L.~Zhang, Y.~Tian, X.~Xia, and X.~Fu, ``Zero-ig: Zero-shot
  illumination-guided joint denoising and adaptive enhancement for low-light
  images,'' in \emph{Proceedings of the IEEE/CVF Conference on Computer Vision
  and Pattern Recognition}, 2024, pp. 3015--3024.

\bibitem{LPIPS}
R.~Zhang, P.~Isola, A.~A. Efros, E.~Shechtman, and O.~Wang, ``The unreasonable
  effectiveness of deep features as a perceptual metric,'' in \emph{Proceedings
  of the IEEE conference on computer vision and pattern recognition}, 2018, pp.
  586--595.

\bibitem{NIQE}
A.~Mittal, R.~Soundararajan, and A.~C. Bovik, ``Making a “completely blind”
  image quality analyzer,'' \emph{IEEE Signal processing letters}, vol.~20,
  no.~3, pp. 209--212, 2012.

\bibitem{EME}
S.~S. Agaian, B.~Silver, and K.~A. Panetta, ``Transform coefficient
  histogram-based image enhancement algorithms using contrast entropy,''
  \emph{IEEE transactions on image processing}, vol.~16, no.~3, pp. 741--758,
  2007.

\bibitem{LOE}
S.~Wang, J.~Zheng, H.-M. Hu, and B.~Li, ``Naturalness preserved enhancement
  algorithm for non-uniform illumination images,'' \emph{IEEE transactions on
  image processing}, vol.~22, no.~9, pp. 3538--3548, 2013.

\end{thebibliography}

\end{document}